\documentclass[a4paper]{article}

\newcommand{\datasetname}{Earnings-21}

\newcommand{\github}{https://github.com/revdotcom/speech-datasets/tree/master/earnings21}
\newcommand{\fstalign}{https://github.com/revdotcom/fstalign}

\newcommand{\fstalignname}{fstalign}

\newcommand{\revkaldi}{\textbf{Kaldi}}
\newcommand{\revend}{\textbf{ESPNet}}
\newcommand{\librispeech}{\textbf{LibriSpeech}}
\newcommand{\asrone}{\textbf{Google}}
\newcommand{\asrtwo}{\textbf{Amazon}}
\newcommand{\asrthree}{\textbf{Microsoft}}
\newcommand{\asrfour}{\textbf{Speechmatics}}

\usepackage{INTERSPEECH2021}
\usepackage{subfiles}
\usepackage{cleveref}
\usepackage{todonotes}
\usepackage{dblfloatfix} 
\usepackage{amsmath,graphicx}
\usepackage{multirow}
\usepackage{subfigure}

\title{\datasetname: A Practical Benchmark for ASR in the Wild}

\name{
Miguel Del Rio$^1$, 
Natalie Delworth$^1$,
Ryan Westerman$^1$,
Michelle Huang$^1$,
Nishchal Bhandari$^1$,
Joseph Palakapilly$^1$,
Quinten McNamara$^1$,
Joshua Dong$^1$,
Piotr {\.Z}elasko$^{2,3}$,
Miguel Jett\'e$^1$
}

\address{
  $^1$Rev.com,\\
  $^2$Center for Language and Speech Processing,
  $^3$Human Language Technology Center of Excellence, Johns Hopkins University, Baltimore, MD, USA}
\email{
miguel.delrio@rev.com
}

\begin{document}

\maketitle

\begin{abstract}
Commonly used speech corpora inadequately challenge academic and commercial ASR systems. In particular, speech corpora lack metadata needed for detailed analysis and WER measurement. In response, we present \textit{\datasetname},
a 39-hour corpus of earnings calls containing entity-dense speech from nine different financial sectors. This corpus is intended to benchmark ASR systems in the wild with special attention towards named entity recognition.
We benchmark four commercial ASR models, two internal models built with open-source tools, and an open-source LibriSpeech model and discuss their differences in performance on \textit{\datasetname}. 
Using our recently released \textit{\fstalignname} tool, we provide a candid analysis of each model's recognition capabilities under different partitions.
Our analysis finds that ASR accuracy for certain NER categories is poor, presenting a significant impediment to transcript comprehension and usage.
\textit{\datasetname} bridges academic and commercial ASR system evaluation and enables further research on entity modeling and WER on real world audio.

\end{abstract}

\section{Introduction}
\label{sec:intro}

Automatic Speech Recognition (ASR) has been adopted for a wide variety of acoustic environments.
Users expect ASR systems to understand a wide range of voices in various settings such as podcasts, quarterly earnings calls, and streaming video captioning.
Whereas there exist multiple techniques that allow to adapt an ASR system to various acoustic conditions~\cite{snyder2015musan,cui2015data,park2019specaugment}, it is also necessary to evaluate the system in its target operating conditions.


At present, common publicly available evaluation sets include LibriSpeech~\cite{librispeech}, Switchboard~\cite{godfrey1992switchboard}, CallHome~\cite{callhome}, Rich Transcription 2007~\cite{rt07}, among others.
The latest of these evaluation sets is over five years old, and none of them feature a wide variety of voices, technical domains, or acoustic environments. Recent corpora such as CHiME-5~\cite{watanabe2020chime} poorly reflect real-world recording conditions.
Furthermore, many of these traditional evaluation sets are not free to use, limiting access to research groups or well-funded
private companies. The most challenging public test suite our team has used in the past is the AMI corpus~\cite{carletta2006announcing}, which features difficult speakers and good variance in recording characteristics.
Recently, \cite{szymanski2020we} have shown that these standard ASR tasks and benchmarks create an overly-optimistic and misleading view of the current state of the art. Whereas the best reported word error rate (WER) results on LibriSpeech (around 1.4\%~\cite{zhang2020pushing}) or Switchboard (around 5\%~\cite{han2018densely}) are encouraging, in reality, most commercially available systems are much closer to 15-20\%~\cite{szymanski2020we} when transcribing user-provided recordings, making ASR a problem that is far from solved.

For transcription services such as Rev, the ASR API is domain agnostic, which necessitates a substantial effort in the procurement of evaluation sets that reflect a wide variety of acoustic environments, domains, voices, and accents. 

In order to bolster the community's efforts in robust ASR research, we release \textit{\datasetname}, an open and free evaluation corpus consisting of earnings call recordings and their corresponding rich transcripts available on Github\footnote{\github}. The main contributions of \textit{\datasetname} are:
\begin{itemize}
    \item A new freely available resource for ASR evaluation, sourced ``in the wild'' from recordings  created during the year 2020
    \item Richly annotated transcripts (with punctuation, true-casing, and named entities) for detailed error analysis
    \item A benchmark of commercial and academic ASR systems on the corpus
    \item \textit{\fstalignname}\footnote{\fstalign}, a novel toolkit for quickly computing WER that leverages NER annotations
\end{itemize}

The rest of the paper is organized as follows: \Cref{sec:dataset} details dataset properties and sourcing methodology, \Cref{sec:results} compares the performance of various ASR systems on our new evaluation set, and \Cref{sec:conclusions} presents our future plans.

\section{The \textit{\datasetname} Dataset}
\label{sec:dataset}

The \textit{\datasetname} dataset consists of 44 public\footnote{Earnings calls fair use legal precedent in \textit{Swatch Group Management Services Ltd. v. Bloomberg L.P.}} earnings calls recorded in 2020 from 9 corporate sectors downloaded from Seeking Alpha\footnote{https://seekingalpha.com/earnings/earnings-call-transcripts}, totalling 39 hours and 15 minutes. Our data selection intends to reflect real world settings with diverse semantic and acoustic properties. 
The files in \textit{\datasetname} contain:
\begin{itemize}
    \item Varied sector-specific technical terminology
    \item A wide range of recording qualities - representative of audio typically received in the wild
    \item Entity-rich transcripts with annotated numerical figures
    \item Semantic and linguistic content unique to the year 2020
\end{itemize}
In particular, earnings calls have vastly varying recording characteristics and speaker profiles in the same call. We do not have any information on this audio metadata other than what can be inferred from the audios themselves. The audios are stored as monaural MP3 files.

To cover a wide range of scenarios common in real-world use cases, we chose recordings that had diverse sample rates as presented in \Cref{tab:sample_rate}. The recordings in this corpus range in length from less than 17 minutes to 1 hour and 34 minutes with the average recording being about 54 minutes in length.

\begin{table}[!h]
    \centering
    \begin{tabular}{|c|c|c|}
         \hline
         \textbf{Sampling rate (Hz)} & \textbf{Recordings} & \textbf{Total time (hh:mm)} \\
         \hline
         44100&7&07:13\\
         24000&21&17:45\\
         22050&5&04:12\\
         16000&6&04:52\\
         11025&5&05:14\\
         \hline
    \end{tabular}
    \caption{Sampling rate distribution across $\text{\datasetname}$ in number of files and total duration.}
    \label{tab:sample_rate}
\end{table}

\begin{figure}[]
  \centering
  \includegraphics[width=0.9\linewidth]{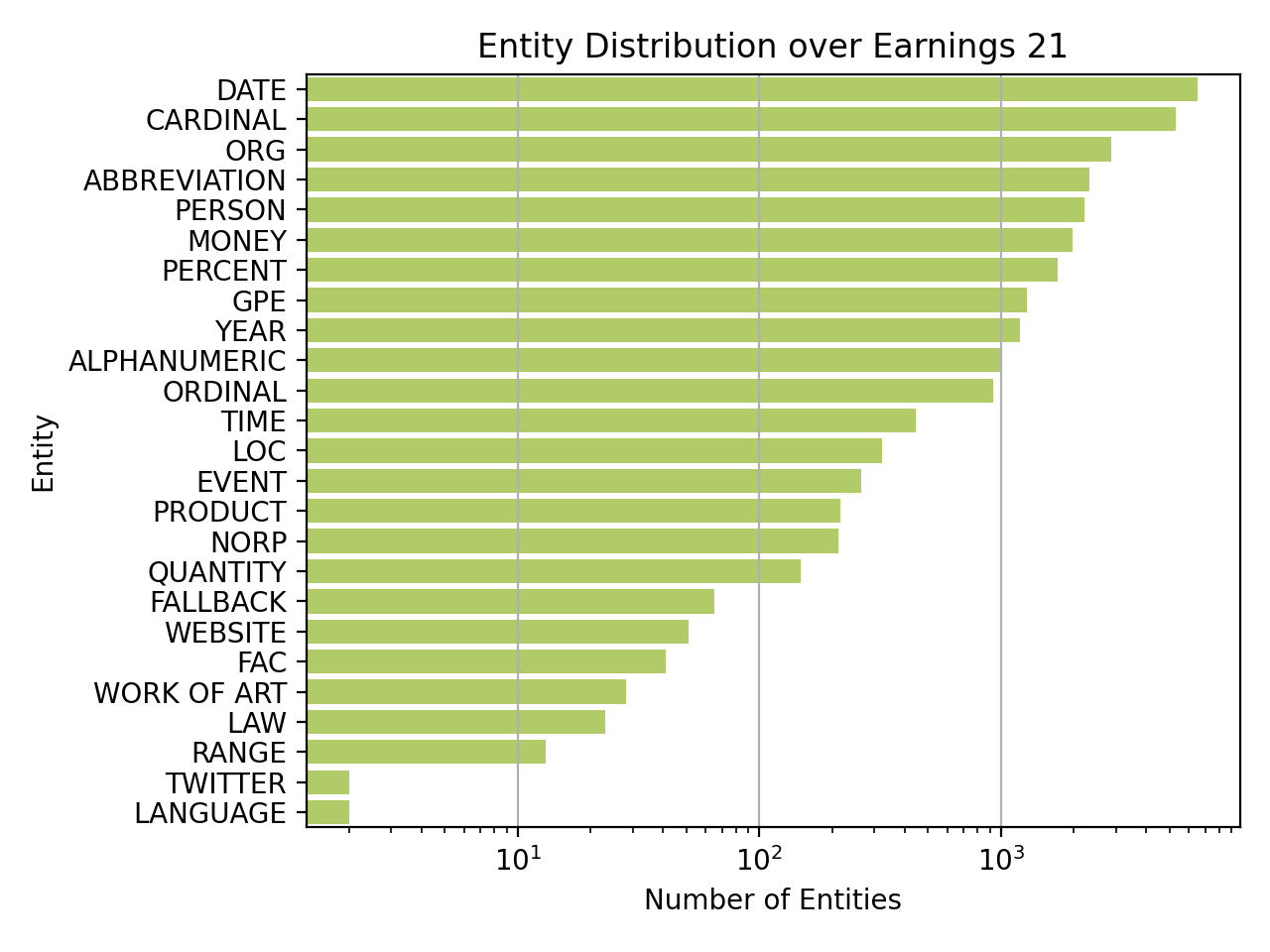}
  \caption{Log-scaled distribution of entities over $\text{\datasetname}$ generated by SpaCy.}
  \label{fig:spacy_entity_distribution}
\end{figure}

\subsection{Earnings call selection}
Seeking Alpha defines 9 different sectors that categorize all earnings calls on their website - these are: Basic Materials, Conglomerates, Consumer Goods, Financial, Healthcare, Industrial Goods, Services, Technology, and Utilities. The average sector has just over 40,000 tokens and has about 4.5 hours of audio. To ensure diverse coverage, we randomly selected 5 calls that occurred in 2020 from each of these sectors.

\subsection{Dataset transcription}
To get accurate transcriptions, we used the Rev.com human transcription service.
These audios were rigorously transcribed by a pool of experienced transcriptionists,
then graded and reviewed by a different pool of senior transcriptionists.
Spot-checking by the paper authors find that transcripts created using this process are highly accurate.

We chose to get ``verbatim'' transcriptions that capture all speech utterances in exactly the same way those words were spoken -- including filler words, false starts, grammatical errors, and other verbal cues or disfluencies\footnote{For more information on Rev.com's verbatim transcription see https://www.rev.com/blog/resources/verbatim-transcription}. We found that verbatim transcripts are more useful for ASR evaluation. Real-world speech features frequent stuttering, repetition, and other disfluencies -- modelling these mistakes is important for accurate transcription of a given recording~\cite{disfphd, heeman1999}.

During transcription, a transcriber found that one of the earnings calls 
contained a large amount of non-English speech; we removed this call\footnote{This leaves the \emph{Conglomerates} sector with only 4 calls.} from the dataset without replacing it because the remaining files still provide adequate coverage of entities over all sectors. 

\subsection{Data preparation}
At Rev, we store reference transcripts in a custom format file we call \texttt{.nlp} files. These files are \texttt{.csv}-inspired pipe-separated (i.e. '$|$') text files that present tokens and their metadata on separate lines.
We assigned NER labels to each transcript in three stages. First we used our internal NER tools to tag tokens that require text normalization such as abbreviations, cardinals, ordinals, and contractions. Next, we applied SpaCy 2.3.5's
NER tags to cover entities our labeller does not tag; these include organizations, people, and nationalities to name a few. Finally, we manually reviewed these tags and updated the entities\footnote{We provide a detailed explanation of the \texttt{nlp} file format and a description of each entity class in our Github.}.  The labeled entities are distributed as shown in \Cref{fig:spacy_entity_distribution}.

As part of this release, we also include all metadata available to us.
In some recordings, speakers are identified by name -- when provided by the transcriptionists we include these in the speaker metadata. 
On a per-file basis, we take advantage of the metadata gathered as part of the data selection. This includes file length in seconds, file sampling rate, the company name / sector, the calls financial quarter, the number of unique speakers, and the total number of utterances in each recording.

\section{Results on \textit{\datasetname}}
\label{sec:results}
We evaluated the transcription accuracy between four commercial ASR systems, two of our own ASR systems, and an open-source Kaldi model on \textit{\datasetname}. 
All models are run using an offline, batch decoding approach (the commercial models are run using their offline API pipeline when available).
Using \textit{\fstalignname}, we provide detailed WER analysis comparison of the earnings call transcription results.
\begin{table*}[ht]
\centering
\begin{subfigure}
    \centering
    \begin{tabular}[t]{|c|c|c|c|c|c|c|c|}
    \hline
        \multirow{2}{*}{Evaluation Set}&\multirow{2}{*}{\asrone}&\multirow{2}{*}{\asrtwo}&\multirow{2}{*}{\asrthree}&\multirow{2}{*}{\asrfour}&\multicolumn{2}{|c|}{\textbf{Rev}}&\textbf{Kaldi.org}\\\cline{6-8}
        &&&&&\revkaldi&\revend&\librispeech\\
        \hline
        \textit{\datasetname}&17.8&17.0&15.8&16.0&13.2&\textbf{11.3}&48.8\\
        \hline
    \end{tabular}
    \caption{Comparison of WER overall models on the
    \textit{\datasetname} dataset.}
    \label{tab:overall}
\end{subfigure}
\begin{subfigure}
    \centering
    \begin{tabular}[t]{|c|c|c|c|c|c|c|c|}
    \hline
        \multirow{2}{*}{Entity}&\multirow{2}{*}{\asrone}&\multirow{2}{*}{\asrtwo}&\multirow{2}{*}{\asrthree}&\multirow{2}{*}{\asrfour}&\multicolumn{2}{|c|}{\textbf{Rev}}&\textbf{Kaldi.org}\\\cline{6-8}
        &&&&&\revkaldi&\revend&\librispeech\\
        \hline
        \textit{Mean Entity}&30.4&28.8&20.7&28.8&19.6&\textbf{16.6}&48.9\\
        \hline
        \multicolumn{8}{|c|}{Easy Entities}\\
        \hline
        
        DATE&9.8&7.8$^\triangledown$&5.0$^\triangledown$&6.5$^\triangledown$&5.5$^\triangledown$&4.6&30.8\\
        TIME&10.0&7.9&6.5&9.0&10.0&5.0&39.3\\
        ORDINAL&7.3$^\triangledown$&8.3&7.6&8.6&8.2&4.3$^\triangledown$&33.4$^\triangledown$\\
        \hline
        \multicolumn{8}{|c|}{Hard Entities}\\
        \hline
        
        FAC&40.7&40.0&34.8&44.1&36.1&36.1&60.2\\
        ORG&35.9&39.9&35.6&44.3&32.5&31.4&68.8\\
        PERSON&48.2$^+$&46.6$^+$&45.2$^+$&51.7$^+$&46.8$^+$&42.1$^+$&75.5$^+$\\
        \hline
    
    \end{tabular}
    \caption{Entity WER on the three of the easiest and three of the hardest entities defined by the overall performance of all models. We denote a model's best and worst performing entities with $^\triangledown$ and $^+$ respectively.}
    \label{tab:entity}
\end{subfigure}
\end{table*}

\subsection{WER calculation}
As ASR systems become more accurate, more sophisticated measurement tools are needed to attenuate the effects of trivial, ambiguous, or otherwise less interesting errors. Our new open-source tool \textit{\fstalignname} enables this by allowing for specific word substitutions and incorporating text normalization information.

To attenuate the effects of trivial errors, our tool uses a curated list of common word transforms that enable synonymous tokens to be leniently substituted. These transforms allow for hypothesis and reference transcripts to differ in semantically insignificant or ambiguous ways without penalizing WER scores. The following are two example transforms:
\begin{center}
    going to $\longrightarrow$ gonna\\
    I'll $\longrightarrow$ I will
\end{center}
In this example, for the reference ``I'm going to win.'', ``I'm gonna win.'' would be a penalty-free hypothesis. In \textit{\datasetname}, these synonym-transformations typically affect 0.3\% of the potential transcript disagreements.

\textit{\fstalignname} also enables custom transformations to specific tokens in the reference set, which is useful for incorporating text normalization information to the WER calculation. This is especially important when benchmarking across commercial providers, as output formats and inverse text normalization capabilities vary widely. As an example, if we have "2021" tagged as a YEAR in the reference, it could have the following normalizations accepted as correct during WER calculation: "twenty twenty one" and "two thousand twenty one". Our data release includes the text normalization information we used for the WER calculations presented here. These transformations affect roughly 5.0\% of reference tokens.

\subsection{Commercial models}
We chose four of the best commercial cloud ASR providers in the market to run our experiments against. For each commercial provider, we selected the model we've found to perform the best in general so that we can get the best possible measurement of where industry stands with respect to our dataset. These providers are {\asrone} (using the Video model), \asrtwo, \asrthree, and \asrfour. These models are all black-box to us and therefore we cannot provide more information about the specifications. The commercial model output is provided in the data release for convenient reproducibility.

\subsection{Internal models}
We trained two models using the popular Kaldi and ESPNet toolkits.
Our models were developed as part of general ASR systems with training data sourced from an
unbiased
selection from our database. Audio is resampled at 16KHz for training and 
inference time.

The first system is a Kaldi~\cite{kaldi} based DNN-HMM trained on 30,000 hours of out-of-domain proprietary audio. The acoustic model is comprised of 80M parameters in interleaved TDNN and LSTM layers~\cite{tdnn_lstm},
 a 3-gram decoder (6M entries), and a 4-gram LM (150M entries) interpolated with a 16M parameter TDNN-LSTM RNNLM for rescoring.


The second system is an ESPnet V2~\cite{watanabe2018espnet} based hybrid CTC/Attention encoder-decoder~\cite{watanabe2017hybrid} model trained on 10,000 hours of out-of-domain proprietary audio. We used the predefined LibriSpeech \textit{conformer7} configuration with 124M parameters, 10,000 BPE~\cite{kudo-richardson-2018-sentencepiece} tokens, and a maximum token length of 8 characters.

For further WER comparison, we also used an open-source Kaldi model\footnote{https://kaldi-asr.org/models/m13} trained on 960 hours from LibriSpeech~\cite{librispeech} with standard 3-way speed perturbation.

\subsection{Comparison}
We present the results of our WER measurements in \Cref{tab:overall}. We find that the $\revend$ model is the most accurate on \textit{\datasetname}. We were surprised to see the chasm between the open-source $\librispeech$ model and the proprietary ASR systems. We posit that this is due to:
(1) domain mismatch due to vastly different acoustic channel and recording characteristics between LibriSpeech and earnings calls, and
(2) orders-of-magnitude difference in amount of data used to train the ASR models, as in the case of our models which used over 10 times the amount of data used to train the open source models.

One recording\footnote{File-id 4346923 in the sector ``Industrial Goods'' and with sampling rate of 16kHz}
showed significantly degraded WER on all models. Manual review reveals that several speakers have heavy accents and low recording quality; many models failed to transcribe large sections of this file.
If this file is excluded from evaluation, our internal $\revkaldi$ model is the most accurate. We have intentionally chosen to keep this difficult file as it presents realistic lens into the variability of audio in the wild.

In the following subsections, we analyze the WER results using different stratifications to better understand the nature of ASR performance on real-world audio.

\begin{table*}
\centering
\begin{subfigure}
    \centering
    \begin{tabular}[t]{|c|c|c|c|c|c|c|c|}
        \hline
        \multirow{2}{*}{Sector}&\multirow{2}{*}{\asrone}&\multirow{2}{*}{\asrtwo}&\multirow{2}{*}{\asrthree}&\multirow{2}{*}{\asrfour}&\multicolumn{2}{|c|}{\textbf{Rev}}&\textbf{Kaldi.org}\\\cline{6-8}
        &&&&&\revkaldi&\revend&\librispeech\\
        \hline
        \textit{Mean Sector}&17.8&17.1&15.8&16.0&13.2&\textbf{11.3}&48.8\\
        \hline
        Conglomerates&15.5$^\triangledown$&15.4$^\triangledown$&14.1$^\triangledown$&14.0$^\triangledown$&8.0$^\triangledown$&10.2&44.1$^\triangledown$\\
        Utilities&15.9&15.9&14.8&14.2&10.3&10.8&45.7\\
        Basic Materials&16.7&15.5&14.6&14.5&11.0&11.1&43.6\\
        Services&16.8&16.6&14.8&15.2&11.5&9.8$^\triangledown$&44.1\\
        Healthcare&17.1&17.1&15.6&16.0&11.0&10.6&44.6\\
        Financial&18.0&17.0&15.6&15.5&13.2&11.5&49.5\\
        Consumer Goods&18.7&17.3&16.0&16.1&12.1&10.3&51.1\\
        Technology&20.6&18.9&17.1&17.4&16.0&12.9&56.3\\
        Industrial Goods&21.2$^+$&20.0$^+$&19.3$^+$&21.0$^+$&25.9$^+$&14.4$^+$&60.2$^+$\\
        
         \hline
    \end{tabular}
    \caption{WER breakdown by sector (domain) defined by Seeking Alpha. We denote a model's best and worst performing sector with $^\triangledown$ and $^+$ respectively.}
    \label{tab:sector}
\end{subfigure}
\begin{subfigure}
    \centering
    \begin{tabular}[t]{|c|c|c|c|c|c|c|c|}
        \hline
        \multirow{2}{*}{Sample Rate (Hz)}&\multirow{2}{*}{\asrone}&\multirow{2}{*}{\asrtwo}&\multirow{2}{*}{\asrthree}&\multirow{2}{*}{\asrfour}&\multicolumn{2}{|c|}{\textbf{Rev}}&\textbf{Kaldi.org}\\\cline{6-8}
        &&&&&\revkaldi&\revend&\librispeech\\
        \hline 
        \textit{Mean Sample Rate}&18.1&17.5&16.2&16.4&14.5&\textbf{11.8}&49.0\\
        \hline
         44100&16.0&15.5$^\triangledown$&14.9&14.4&10.0&9.6$^\triangledown$&40.5$^\triangledown$\\
         24000&17.3&16.3&15.0&15.2&11.3&10.4&49.7\\
         22050&14.6$^\triangledown$&15.6&13.4$^\triangledown$&12.6$^\triangledown$&8.9$^\triangledown$&10.5&43.3\\
         16000&22.9$^+$&21.1$^+$&20.4$^+$&22.3$^+$&28.1$^+$&15.1$^+$&59.5$^+$\\
         11025&19.9&19.1&17.2&17.5&14.2&13.5&52.2\\
        \hline
    \end{tabular}
    \caption{WER breakdown by recording's sampling rate. We denote a model's best and worst performing sample rate with $^\triangledown$ and $^+$ respectively. See \Cref{tab:sample_rate} for information on the distribution of sample rates.}
    \label{tab:sr}
\end{subfigure}
\end{table*}

\subsubsection{Entity recognition}
We show how different models performed on the identified named entity classes in \Cref{tab:entity}. 
Looking for the entity classes with the lowest WER across all models, we find  \emph{DATE}, \emph{ORDINAL}, and \emph{TIME} are the easiest entity classes to recognize. We hypothesize that these entities are easier due to their structured pattern and frequent appearance in training data. On the other hand, looking at the entities with the highest overall WER, we find that \emph{FAC}, \emph{ORG}, and \emph{PERSON} are difficult, which may be due to higher lexical diversity in those categories.
ASR systems incorporate language models which are particularly sensitive to sparsity, making recognition of rare or novel token sequences difficult without special modeling.

\subsubsection{Domain}
We demonstrate the WER results with topic-domain partitioning as defined by business sector in \Cref{tab:sector}. The data set shows poorest accuracy in \emph{Industrial Goods} and \emph{Technology} sectors and best accuracy in the \emph{Conglomerates} sector. The measured difficulty of transcribing \emph{Industrial Goods} is attributed to the most difficult file in the corpus; treating that file as an outlier leads us to believe the sector's difficulty is average.
In the \emph{Technology} domain, models suffer from large contiguous deletions which can account for over 40\% of the errors in a file.
More work needs to be done to understand the discrepancies in accuracy between sectors.

\subsubsection{Sampling rate}
The data set recordings have diverse sampling rates. We compare ASR performance with respect to sampling rate in \Cref{tab:sr}. 
We find that most systems perform similarly at 22050Hz and greater sampling rates. We note that our most difficult file to transcribe is 16kHz and skews the WER averages, but omitting this file shows a clear positive correlation between sampling rate and accuracy. 11025Hz is a particularly difficult sampling rate, likely due to limited bandwidth.

We conclude that higher sampling rate generally leads to more accurate transcription. Audios with higher sampling rates provide higher quality speech signals even after downsampling. Even though commercial ASR systems are black-box systems for which we do not know what sampling rate was used for their input data, it is likely that models are trained with high sampling rate recordings, which may explain the association with WER.

\section{Conclusion}
\label{sec:conclusions}
We show that there still exist major obstacles to speech recognition in the wild. With our data set release, we challenge researchers to deal with real-world audio. We also provide \textit{\fstalignname} as a tool to enable the research community to focus on higher-importance entity WER and move past trivial errors.

We will continue to improve the metadata for our \textit{\datasetname} corpus and invite others to contribute as well. We hope this release is the first of many towards providing a realistic view of speech in the wild.
We encourage industry leaders and academic researchers to continue research in this vein,
as continued efforts towards modeling real-world challenges and up-to-date data will be the future of ASR.

\section{Acknowledgements}
We would like to thank the transcriptionists and contractors who dedicated hundreds of hours and helped us to create and improve the quality of this dataset.

\bibliographystyle{IEEEtran}
\bibliography{main}

\begin{thebibliography}{10}
\providecommand{\url}[1]{#1}
\csname url@samestyle\endcsname
\providecommand{\newblock}{\relax}
\providecommand{\bibinfo}[2]{#2}
\providecommand{\BIBentrySTDinterwordspacing}{\spaceskip=0pt\relax}
\providecommand{\BIBentryALTinterwordstretchfactor}{4}
\providecommand{\BIBentryALTinterwordspacing}{\spaceskip=\fontdimen2\font plus
\BIBentryALTinterwordstretchfactor\fontdimen3\font minus
  \fontdimen4\font\relax}
\providecommand{\BIBforeignlanguage}[2]{{%
\expandafter\ifx\csname l@#1\endcsname\relax
\typeout{** WARNING: IEEEtran.bst: No hyphenation pattern has been}%
\typeout{** loaded for the language `#1'. Using the pattern for}%
\typeout{** the default language instead.}%
\else
\language=\csname l@#1\endcsname
\fi
#2}}
\providecommand{\BIBdecl}{\relax}
\BIBdecl

\bibitem{snyder2015musan}
D.~Snyder, G.~Chen, and D.~Povey, ``Musan: A music, speech, and noise corpus,''
  \emph{arXiv preprint arXiv:1510.08484}, 2015.

\bibitem{cui2015data}
X.~Cui, V.~Goel, and B.~Kingsbury, ``Data augmentation for deep neural network
  acoustic modeling,'' \emph{IEEE/ACM Transactions on Audio, Speech, and
  Language Processing}, vol.~23, no.~9, pp. 1469--1477, 2015.

\bibitem{park2019specaugment}
D.~S. Park, W.~Chan, Y.~Zhang, C.-C. Chiu, B.~Zoph, E.~D. Cubuk, and Q.~V. Le,
  ``Specaugment: A simple data augmentation method for automatic speech
  recognition,'' \emph{arXiv preprint arXiv:1904.08779}, 2019.

\bibitem{librispeech}
V.~Panayotov, G.~Chen, D.~Povey, and S.~Khudanpur, ``Librispeech: an asr corpus
  based on public domain audio books,'' in \emph{2015 IEEE international
  conference on acoustics, speech and signal processing (ICASSP)}.\hskip 1em
  plus 0.5em minus 0.4em\relax IEEE, 2015, pp. 5206--5210.

\bibitem{godfrey1992switchboard}
J.~J. Godfrey, E.~C. Holliman, and J.~McDaniel, ``Switchboard: Telephone speech
  corpus for research and development,'' in \emph{Acoustics, Speech, and Signal
  Processing, IEEE International Conference on}, vol.~1.\hskip 1em plus 0.5em
  minus 0.4em\relax IEEE Computer Society, 1992, pp. 517--520.

\bibitem{callhome}
C.~Cieri, D.~Miller, and K.~Walker, ``The fisher corpus: a resource for the
  next generations of speech-to-text.'' in \emph{LREC}, vol.~4, 2004, pp.
  69--71.

\bibitem{rt07}
J.~Fiscus, J.~Ajot, and J.~Garofolo, ``\BIBforeignlanguage{en}{The rich
  transcription 2007 meeting recognition evaluation}.''\hskip 1em plus 0.5em
  minus 0.4em\relax The Joint Proceedings of the 2006 CLEAR and RT Evaluations,
  2007.

\bibitem{watanabe2020chime}
S.~Watanabe, M.~Mandel, J.~Barker, E.~Vincent, A.~Arora, X.~Chang,
  S.~Khudanpur, V.~Manohar, D.~Povey, D.~Raj \emph{et~al.}, ``Chime-6
  challenge: Tackling multispeaker speech recognition for unsegmented
  recordings,'' \emph{arXiv preprint arXiv:2004.09249}, 2020.

\bibitem{carletta2006announcing}
J.~Carletta, ``Announcing the ami meeting corpus,'' \emph{The ELRA Newsletter},
  vol.~11, no.~1, pp. 3--5, 2006.

\bibitem{szymanski2020we}
\BIBentryALTinterwordspacing
P.~Szyma{\'n}ski, P.~{\.Z}elasko, M.~Morzy, A.~Szymczak, M.~{\.Z}y{\l}a-Hoppe,
  J.~Banaszczak, L.~Augustyniak, J.~Mizgajski, and Y.~Carmiel, ``{WER} we are
  and {WER} we think we are,'' in \emph{Findings of the Association for
  Computational Linguistics: EMNLP 2020}.\hskip 1em plus 0.5em minus
  0.4em\relax Online: Association for Computational Linguistics, Nov. 2020, pp.
  3290--3295. [Online]. Available:
  \url{https://www.aclweb.org/anthology/2020.findings-emnlp.295}
\BIBentrySTDinterwordspacing

\bibitem{zhang2020pushing}
Y.~Zhang, J.~Qin, D.~S. Park, W.~Han, C.-C. Chiu, R.~Pang, Q.~V. Le, and Y.~Wu,
  ``Pushing the limits of semi-supervised learning for automatic speech
  recognition,'' \emph{arXiv preprint arXiv:2010.10504}, 2020.

\bibitem{han2018densely}
K.~J. Han, A.~Chandrashekaran, J.~Kim, and I.~Lane, ``Densely connected
  networks for conversational speech recognition.'' in \emph{INTERSPEECH},
  2018, pp. 796--800.

\bibitem{disfphd}
E.~E. Shriberg, ``{Preliminaries to a theory of speech disfluencies},'' 1994.

\bibitem{heeman1999}
P.~A. Heeman and J.~Allen, ``{Speech repains, intonational phrases, and
  discourse markers: modeling speakers’ utterances in spoken dialogue},''
  1999.

\bibitem{kaldi}
D.~Povey, A.~Ghoshal, G.~Boulianne, L.~Burget, O.~Glembek, N.~Goel,
  M.~Hannemann, P.~Motlicek, Y.~Qian, P.~Schwarz \emph{et~al.}, ``The kaldi
  speech recognition toolkit,'' in \emph{IEEE 2011 workshop on automatic speech
  recognition and understanding}, no. CONF.\hskip 1em plus 0.5em minus
  0.4em\relax IEEE Signal Processing Society, 2011.

\bibitem{tdnn_lstm}
V.~Peddinti, Y.~Wang, D.~Povey, and S.~Khudanpur, ``Low latency acoustic
  modeling using temporal convolution and lstms,'' vol.~25, no.~3.\hskip 1em
  plus 0.5em minus 0.4em\relax IEEE, 2017, pp. 373--377.

\bibitem{watanabe2018espnet}
\BIBentryALTinterwordspacing
S.~Watanabe, T.~Hori, S.~Karita, T.~Hayashi, J.~Nishitoba, Y.~Unno, N.~{Enrique
  Yalta Soplin}, J.~Heymann, M.~Wiesner, N.~Chen, A.~Renduchintala, and
  T.~Ochiai, ``{ESPnet}: End-to-end speech processing toolkit,'' in
  \emph{Proceedings of Interspeech}, 2018, pp. 2207--2211. [Online]. Available:
  \url{http://dx.doi.org/10.21437/Interspeech.2018-1456}
\BIBentrySTDinterwordspacing

\bibitem{watanabe2017hybrid}
S.~Watanabe, T.~Hori, S.~Kim, J.~R. Hershey, and T.~Hayashi, ``Hybrid
  ctc/attention architecture for end-to-end speech recognition,'' \emph{IEEE
  Journal of Selected Topics in Signal Processing}, vol.~11, no.~8, pp.
  1240--1253, 2017.

\bibitem{kudo-richardson-2018-sentencepiece}
\BIBentryALTinterwordspacing
T.~Kudo and J.~Richardson, ``{S}entence{P}iece: A simple and language
  independent subword tokenizer and detokenizer for neural text processing,''
  in \emph{Proceedings of the 2018 Conference on Empirical Methods in Natural
  Language Processing: System Demonstrations}.\hskip 1em plus 0.5em minus
  0.4em\relax Brussels, Belgium: Association for Computational Linguistics,
  Nov. 2018, pp. 66--71. [Online]. Available:
  \url{https://www.aclweb.org/anthology/D18-2012}
\BIBentrySTDinterwordspacing

\end{thebibliography}

\end{document}